\documentclass{article}

\PassOptionsToPackage{numbers, compress}{natbib}

     \usepackage[final]{neurips_2019}

\usepackage[utf8]{inputenc} 
\usepackage[T1]{fontenc}    
\usepackage{hyperref}       
\usepackage{url}            
\usepackage{booktabs}       
\usepackage{amsfonts}       
\usepackage{nicefrac}       
\usepackage{microtype}      
\usepackage{graphicx}
\usepackage{multirow}
\usepackage{amsmath,amsthm, amssymb, latexsym}
\usepackage{array}
\usepackage{url}
\usepackage{xcolor}
\usepackage{multirow}
\usepackage{float}
\usepackage{wrapfig}

\title{Structured Sparsification of \\Gated Recurrent Neural Networks}

\author{
  Ekaterina Lobacheva$\bf{}^{1}$\thanks{Equal  contribution}\;,\, Nadezhda Chirkova$\bf{}^{1}$\footnotemark[1]\;,\, Alexander Markovich$\bf{}^{2}$,\, Dmitry Vetrov$\bf{}^{1,3}$\\
  ${}^1$Samsung-HSE Laboratory, National Research University Higher School of Economics\\
  ${}^2$National Research University Higher School of Economics\\
  ${}^3$Samsung AI Center 
	 \hspace{35pt} Moscow, Russia\\
  \texttt{\{elobacheva, nchirkova, dvetrov\}@hse.ru, amarkovich@edu.hse.ru}\\}

\begin{document}

\maketitle

\begin{abstract}
  Recently, a lot of techniques were developed to sparsify the weights of neural networks and to remove networks' structure units, e.\,g.\, neurons. We adjust the existing sparsification approaches to the gated recurrent architectures. Specifically, in addition to the sparsification of 
weights and neurons, we propose sparsifying the preactivations of gates. This makes some gates constant and simplifies LSTM structure. We test our approach on the text classification and language modeling tasks. We observe that the resulting structure of gate sparsity depends on the task and connect the learned structure to the specifics of the particular tasks. Our method also improves neuron-wise compression of the model in most of the tasks.
\end{abstract}

\section{Introduction}
Recurrent neural networks (RNNs) yield high-quality results in many applications
but often are memory- and time-consuming due to a large number of parameters.
A popular approach for RNN compression is sparsification (setting a lot of weights to zero), it may compress RNN orders of times with only a slight quality drop or even with quality improvement due to the regularization effect~\cite{pruning}.

\begin{figure}[H]
    \centering
        \begin{tabular}{p{4cm}p{10cm}}
           \multirow{2}{*}{\includegraphics[height=3.1cm]{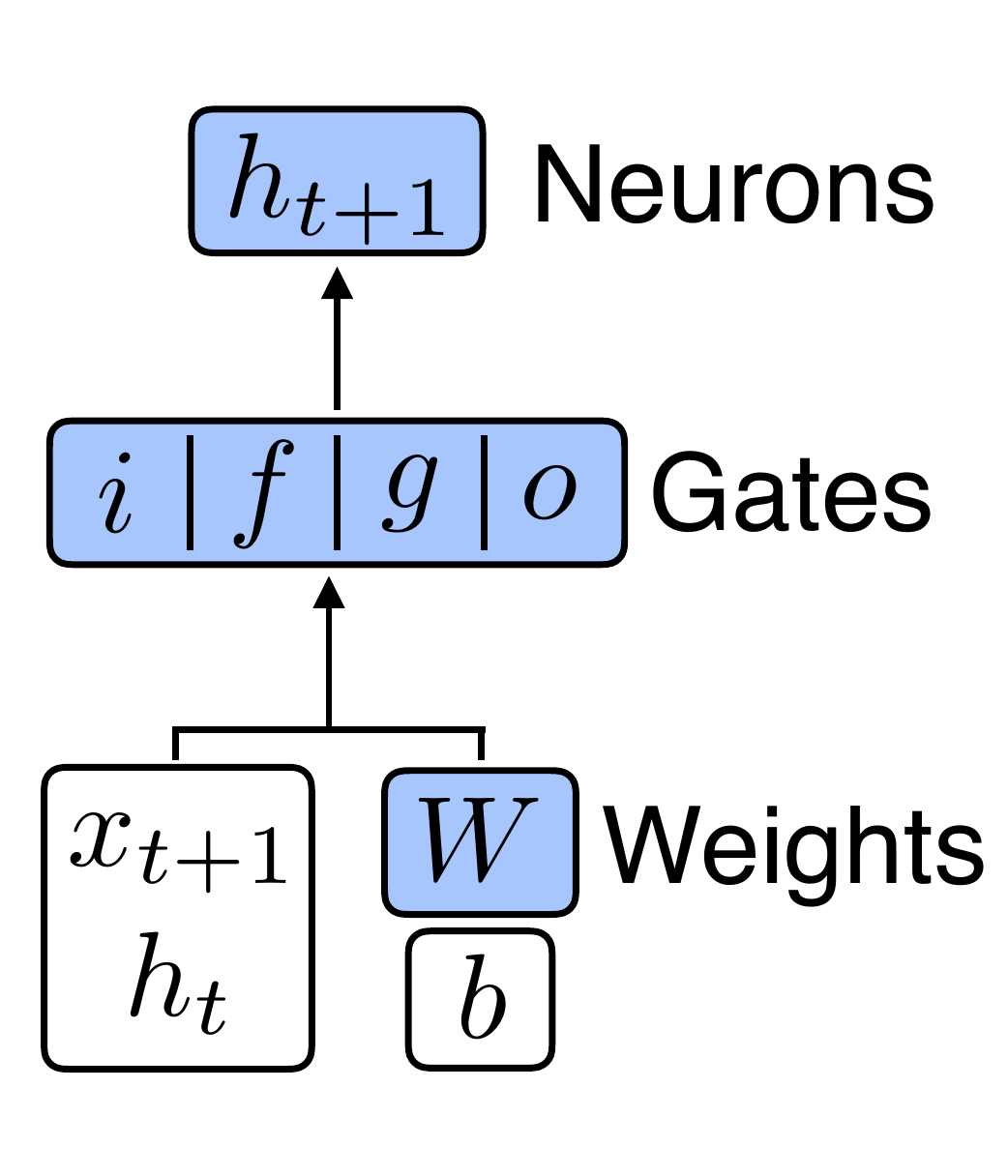}}&
           \raisebox{-0.85\totalheight}{
           \includegraphics[height=1.4cm]{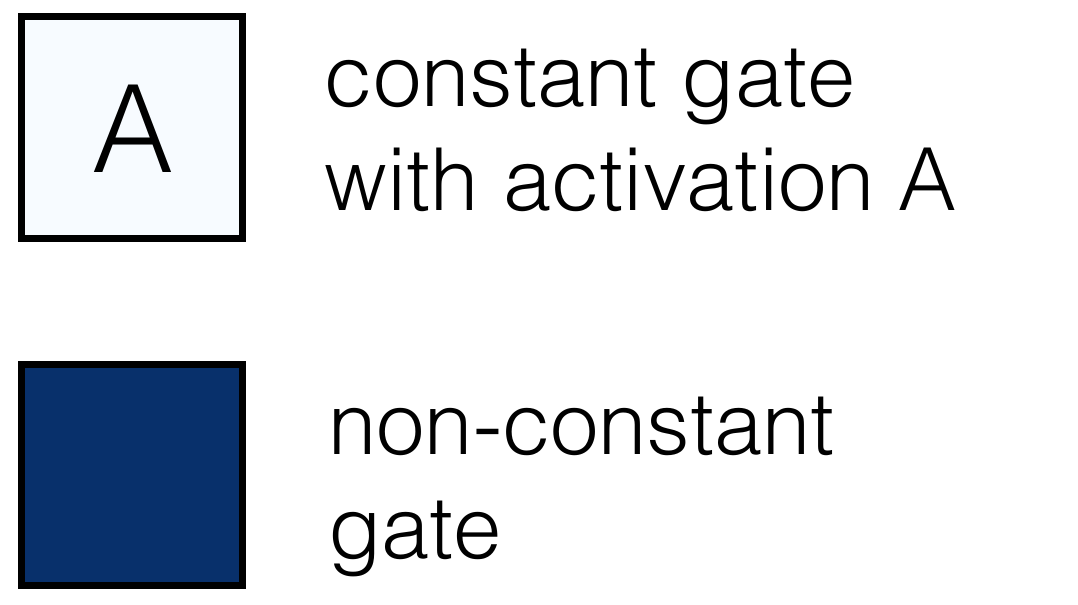}
           \includegraphics[height=1.6cm]{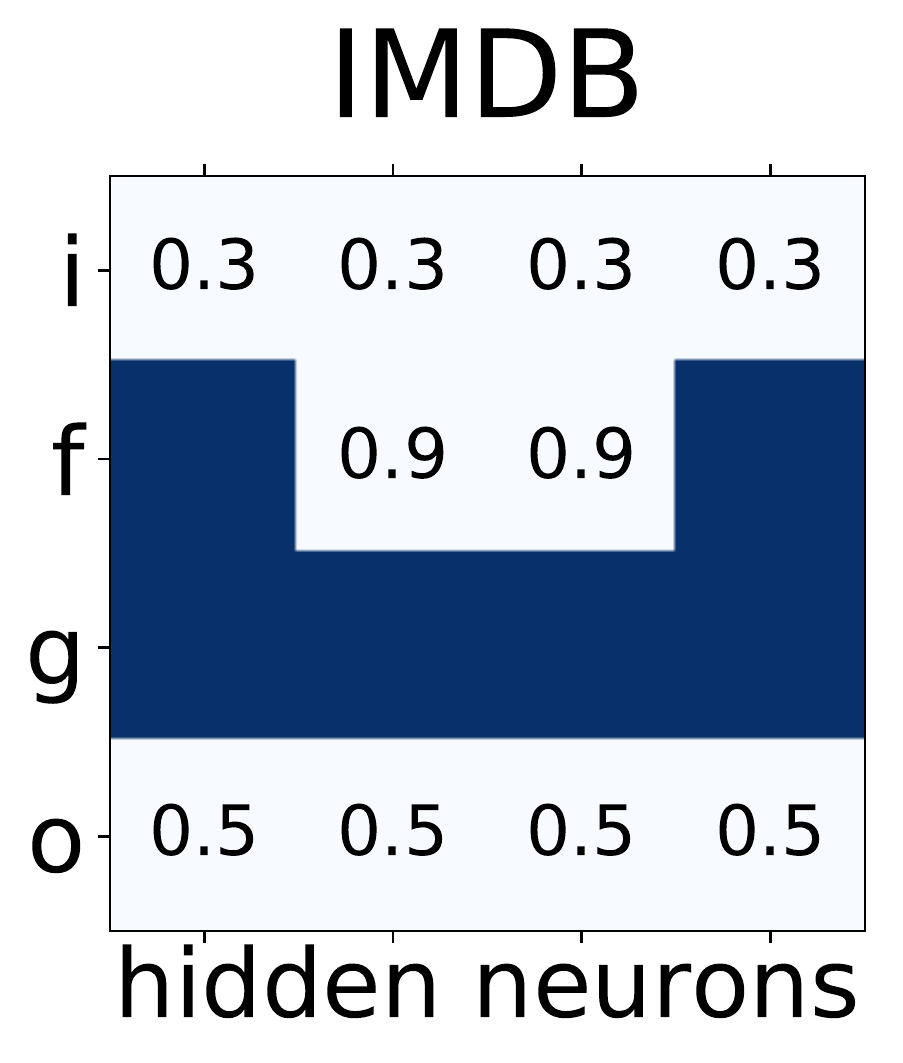} \hspace{0.2cm}
           \includegraphics[height=1.6cm]{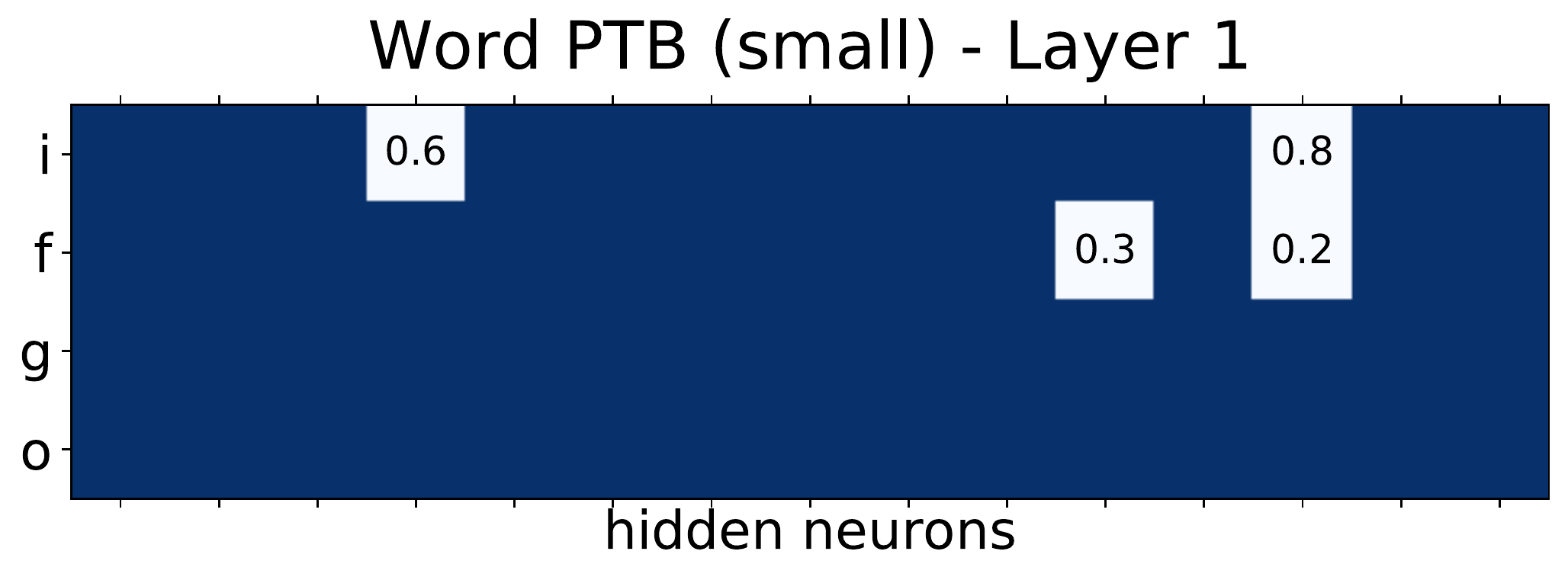}}\\
             &\includegraphics[height=1.6cm]{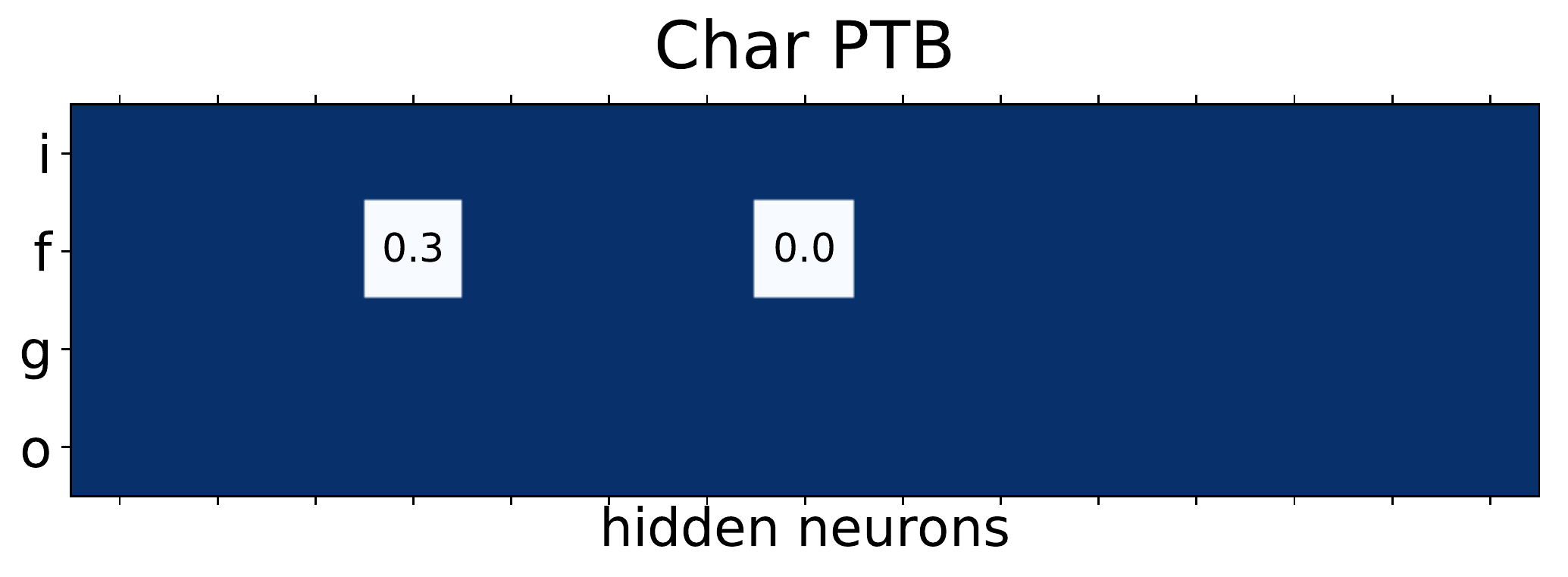}
             \includegraphics[height=1.6cm]{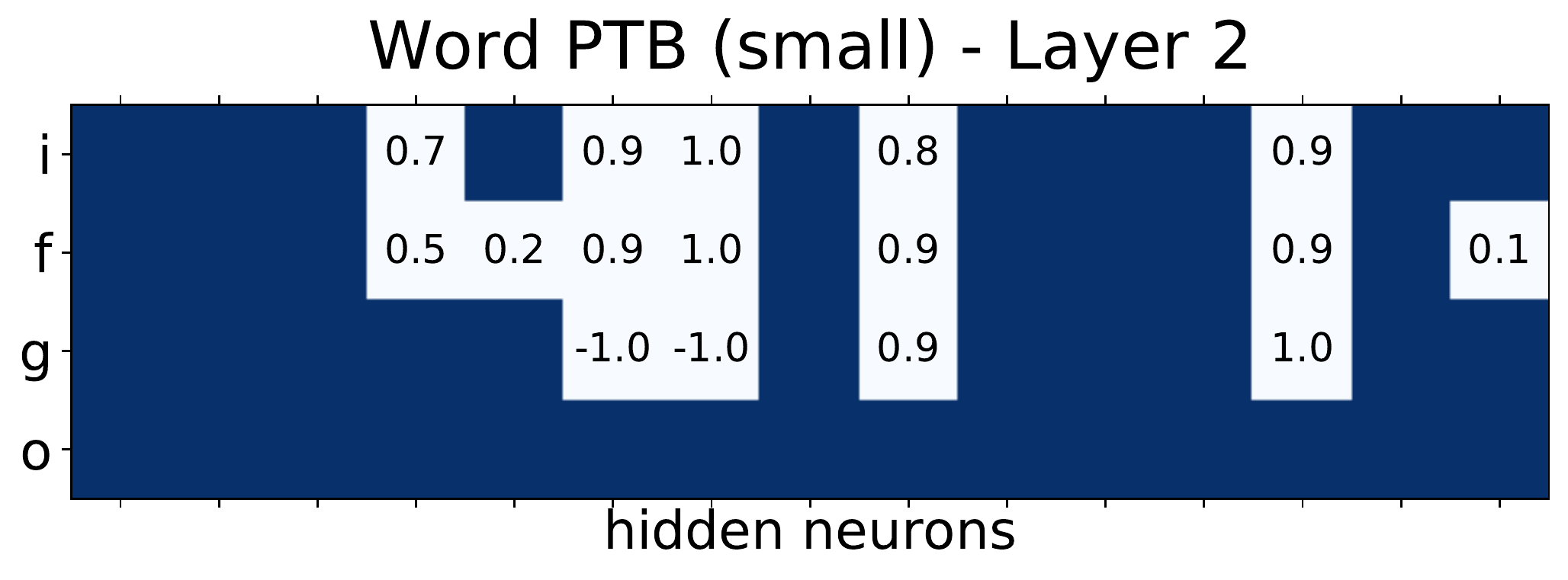}  \\
        \end{tabular}
        \caption{Left: proposed sparsification scheme for LSTM with three levels of sparsity (shown in blue). Right: the resulting gate structures obtained with the proposed modification of the Bayesian sparsification approach (Bayes W+G+N). For PTB, only 15 randomly chosen neurons are presented. 
        }
        \label{fig:idea}
\end{figure}

Sparsification of the RNN is usually performed either at the level of individual weights (unstructured sparsification)~\cite{intel,pruning,emnlp}  or at the level of neurons~\cite{groupsparseLSTM} (structured sparsification --- removing weights by groups corresponding to neurons). The latter additionally accelerates the testing stage. However, most of the modern recurrent architectures (e.\,g.\ LSTM~\cite{lstm} or GRU~\cite{gru}) have a gated structure. We propose to add an 
intermediate level of sparsification between individual weights~\cite{emnlp} and neurons~\cite{groupsparseLSTM} ---
gates (see fig.~\ref{fig:idea}, left). 
Precisely, we remove weights by groups corresponding to gates, which makes{\parfillskip=0pt
\parskip=0pt
\par}
\noindent some gates constant, independent of the inputs, and equal to the activation function of the bias.
As a  result, the LSTM/GRU structure is simplified.
With this intermediate level introduced, we obtain a three-level sparsification hierarchy: sparsification of individual weights helps to sparsify gates (make them constant), and sparsification of gates helps to sparsify neurons (remove them from the model). 

The described idea can be implemented 
for any gated architecture in any sparsification framework. 
We implement the idea for LSTM in two frameworks: pruning~\cite{groupsparseLSTM} and Bayesian sparsification~\cite{emnlp}
and observe that resulting gate structures (which gates are constant and which are not) vary for different NLP tasks. We analyze these gate structures and connect them 
to the specifics of the particular tasks. The proposed method also improves neuron-wise compression of the RNN in most cases.

\section{Proposed method}
\subsection{Main idea\label{idea}}

\begin{wrapfigure}[14]{r}{.5\textwidth} 
\vspace{-4.8ex}
\centering
\includegraphics[height=3.2cm]{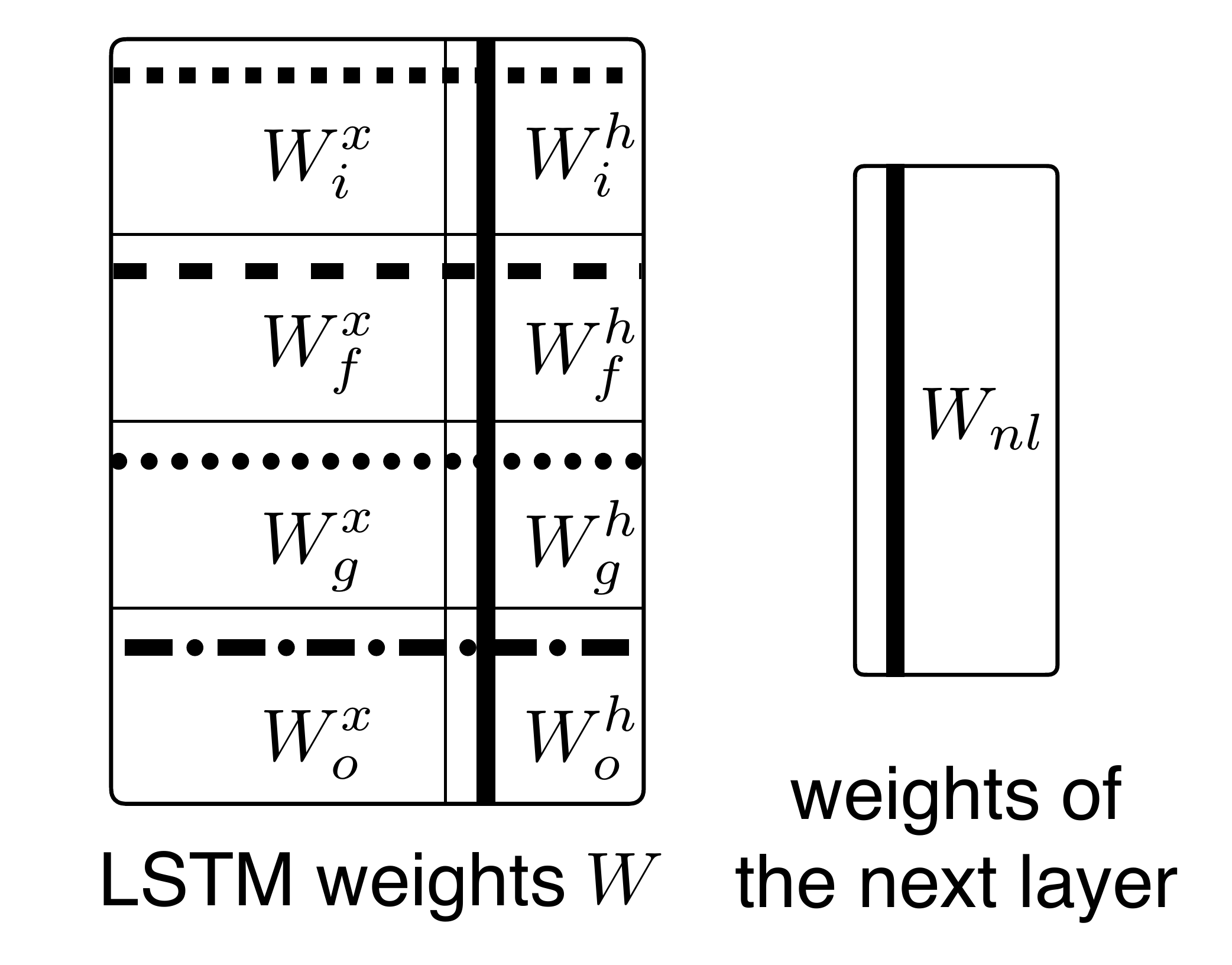} 
\caption{Proposed sparsification scheme for LSTM. Groups of weights for four gates (horizontal lines) and a neuron (vertical lines).}
\label{fig:words}
\end{wrapfigure}

In this section, we describe the three-level sparsification approach for LSTM. LSTM cell is composed of input, forget and output gates ($i$, $f$, $o$) and information flow $g$ 
(which we also call gate for brevity). All four gates are computed in a similar way, for example, for the input gate:
\[
i = sigm(W^x_i x_{t} + W^h_i h_{t-1}+b_i).
\]
To make a gate constant, we need to zero out a corresponding row of the LSTM weight matrix $W$ (see dotted horizontal lines in fig.~\ref{fig:words}).
We do not sparsify biases because they do not take up much memory compared to the weight matrices. 
For example, if we set the $k$-th row of matrices $W^x_i$ and $W^h_i$ to zero, there are no ingoing connections to the corresponding gate, so the $k$-th input gate becomes constant, independent of $x_t$ and $h_{t-1}$ and equal to $sigm(b_{i, k})$.  As a result, we do not need to compute the $k$-th input gate on a forward pass and can use a precomputed value.
We can construct the mask (whether the gate is constant or not) and use it to insert constant values into gate vectors $i, f, g, o$. This lessens the amount of computations on the forward pass.

To remove a neuron, we need to zero out a corresponding column of the LSTM weight matrix $W$ and of the next layer matrix (see solid vertical lines in fig.~\ref{fig:words}). 
This ensures that there are no outgoing connections from the neuron, and the neuron does not affect the network output. 

To sum up, our three-level hierarchy of gated RNN sparsification works as follows. Ideally, our goal is to remove a hidden neuron, this leads to the most effective compression and acceleration. If we don't remove the hidden neuron, some of its four gates may become constant; this also saves computation and memory. If some gate is still non-constant, some of its weights may become zero; this reduces the size of the model.

\subsection{Implementation of the idea}
We incorporate the proposed intermediate level of sparsification into two sparsification frameworks (more details are given in Appendices~\ref{A} and~\ref{B}):

{\bf Pruning.} We apply Lasso to individual weights and group Lasso~\cite{group_lasso} to five groups of the LSTM weights (four gate groups and one neuron group, see fig.~\ref{fig:words}). We use the same pruning algorithm as in Intrinsic Sparse Structure (ISS)~\cite{groupsparseLSTM}, a structured pruning approach developed specifically for LSTM. In contrast to our approach, 
    they do not sparsify gates, and remove a neuron if all its ingoing and outgoing connections are set to zero.

{\bf Bayesian sparsification.} We rely on Sparse Variational Dropout~\cite{dmolch, emnlp} to sparsify individual weights. Following \cite{chris}, for each neuron, we introduce a group weight which is multiplied by the output of this neuron in the computational graph (setting to zero this group weight entails removing the neuron). To sparsify gates, for each gate we introduce a separate group weight which is multiplied by the preactivation of the gate before adding a bias (setting to zero this group weight makes the gate constant).

\section{Experiments}

\begin{table*}[ht!]
  \normalsize
  \centering
  \begin{tabular}{clcccc}
Task & Method & Quality & Compr.  & Neurons & Gates\\ 
\hline
IMDb &\,Original  & \textbf{84.1} & 1x & $128$ & 512 \\
\multicolumn{1}{c}{\textcolor{gray}{Accuracy}\!\!}
&\,Bayes W+N~\cite{emnlp}+\cite{chris}  & 83.98 & 17874x & $5$ & $12$\\ 
\multicolumn{1}{c}{\textcolor{gray}{\%}\!\!}&\,Bayes W+G+N  & 83.98 & {\bf19747x} & $\bf 4$ & {\bf 6}\\ 
\hline
AGNews &\,Original  & {\bf 90.6} & 1x & $ 512$& $2048$\\
\multicolumn{1}{c}{\textcolor{gray}{Accuracy}\!\!}
&\,Bayes W+N~\cite{emnlp}+\cite{chris}  & 88.55 & 645x & $17$ &  $62$ \\ 
\multicolumn{1}{c}{\textcolor{gray}{\%}\!\!}&\,Bayes W+G+N  & 88.41 & {\bf647x}  & $\bf 14$  & {\bf39}\\ 
\hline
Char PTB &\,Original  & $1.499 - 1.454$ & 1x & $ 1000$ &4000\\
\multicolumn{1}{c}{\textcolor{gray}{Bits-per}\!\!}
&\,Bayes W+N~\cite{emnlp}+\cite{chris}  & $1.478 -  1.430$ & \bf10.2x & $\bf 390$& {\bf 1560}\\ 
\multicolumn{1}{c}{\textcolor{gray}{-char}\!\!}&\,Bayes W+G+N & $\bf 1.467 -  1.425$ & 9.8x & $404$& 1563\\
\hline
&\,Original  & 120.28 -- 114.41 & 1x & $200 - 200$& 800 -- 800\\
Word PTB
&\,Bayes W+N~\cite{emnlp}+\cite{chris}  & 110.25 -- 104.81  & \textbf{11.65x}  & 68 -- 110 & 272 -- 392 \\ 
(small)&\,Bayes W+G+N  & \textbf{109.98  -- 104.45}  & 11.44x  & \textbf{52 -- 108} & \textbf{197 -- 349} \\
\cline{2-6}
\multicolumn{1}{c}{\textcolor{gray}{Perplexity}\!\!}
&\,Prun. W+N~\cite{groupsparseLSTM} & 110.34 -- 106.25 & 1.44x & 72 -- 123 & 288 -- 492   \\ 
&\,Prun. W+G+N & \textbf{110.04 -- 105.64} & \textbf{1.49x}  & \textbf{64 -- 115}  & \textbf{193 -- 442} \\	
\hline
Word PTB&\,Original  & 82.57 -- 78.57 & 1x &  $1500 - 1500$ & $6000 - 6000$ \\
(large)&\,Prun. W+N~\cite{groupsparseLSTM}  & \textbf{81.25 -- 77.62} & 2.97x & 324 -- 394 & 1296 -- 1576\\ 
\multicolumn{1}{c}{\textcolor{gray}{Perplexity}\!\!}&\,Prun. W+G+N  & \textbf{81.24 -- 77.82} & \textbf{3.22x} & \textbf{252 -- 394} & \textbf{881 -- 1418} \\ 	
\end{tabular}
\caption{Quantitative results.  For PTB, we evaluate quality on validation and test sets. Compression is equal to $|W|/|W\neq0|$.
In the last  columns, numbers of remaining hidden neurons and non-constant gates are given.
\cite{emnlp}+\cite{chris} means modified approach~\cite{emnlp} with additional group weights for neurons~\cite{chris}.}\label{tab:pruning}
\end{table*}

In the pruning framework,
we perform experiments on word-level language modeling (LM) on a PTB dataset~\cite{ptb} following ISS~\citep{groupsparseLSTM}. We use a standard model of~\citet{zaremba14} of two sizes (small and large) with an embedding layer, two LSTM layers, and a fully-connected output layer (Emb + 2 LSTM + FC). Here regularization is applied only to LSTM layers following~\cite{groupsparseLSTM}, and its strength is selected using grid search so that qualities of ISS and our model are approximately equal.

In the Bayesian framework, we perform an evaluation on the
text classification (datasets IMDb~\cite{IMDB} and AGNews~\cite{agnews}) and language modeling (dataset PTB, character and word level tasks) following~\cite{emnlp}. The architecture for the character-level LM is LSTM + FC, for the text classification is Emb + LSTM + FC on the last hidden state, for the word level LM is the same as in pruning. Here we regularize and sparsify all layers following~\cite{emnlp}.

Sizes of LSTM layers may be found in tab.~\ref{tab:pruning}. Embedding layers have 300/200/1500 neurons for classification tasks/small/large word level LM. 
More experimental details are given in Appendix~\ref{setup}. 

\subsection{Quantitative results}
We compare our three-level sparsification approach (W+G+N) with the original dense model and a two-level sparsification (weights and neurons, W+N) in tab.~\ref{tab:pruning}. 
We do not compare two frameworks between each other; our goal is to show that the proposed idea improves results in both frameworks. 

In most experiments, our method improves gate-wise and neuron-wise compression of the model without a quality drop. The only exception is the character-level LM, which we discuss later. The numbers for compression are not comparable between two frameworks because in pruning only LSTM layers are sparsified while in the Bayesian framework all layers in the network are sparsified.

\subsection{Qualitative results}

Below we analyze the resulting gate structure for different tasks, models and sparsification approaches.

{\bf Gate structure depends on the task.} 
Figure~\ref{fig:idea}, right shows the typical examples of the gate structures of the remaining hidden neurons obtained using the Bayesian approach. We observe that the gate structure varies for different tasks. For the word-level LM task, output gates are very important because models need both store all the information about the input in the memory and output only the current prediction at each timestep. On the contrary, for text classification tasks, models need to output the answer only once at the end of the sequence, hence they rarely use output gates. The{\parfillskip=0pt
\parskip=0pt
\par}
\begin{wrapfigure}[14]{r}{.49\textwidth} 
\centering
\includegraphics[height=3.2cm]{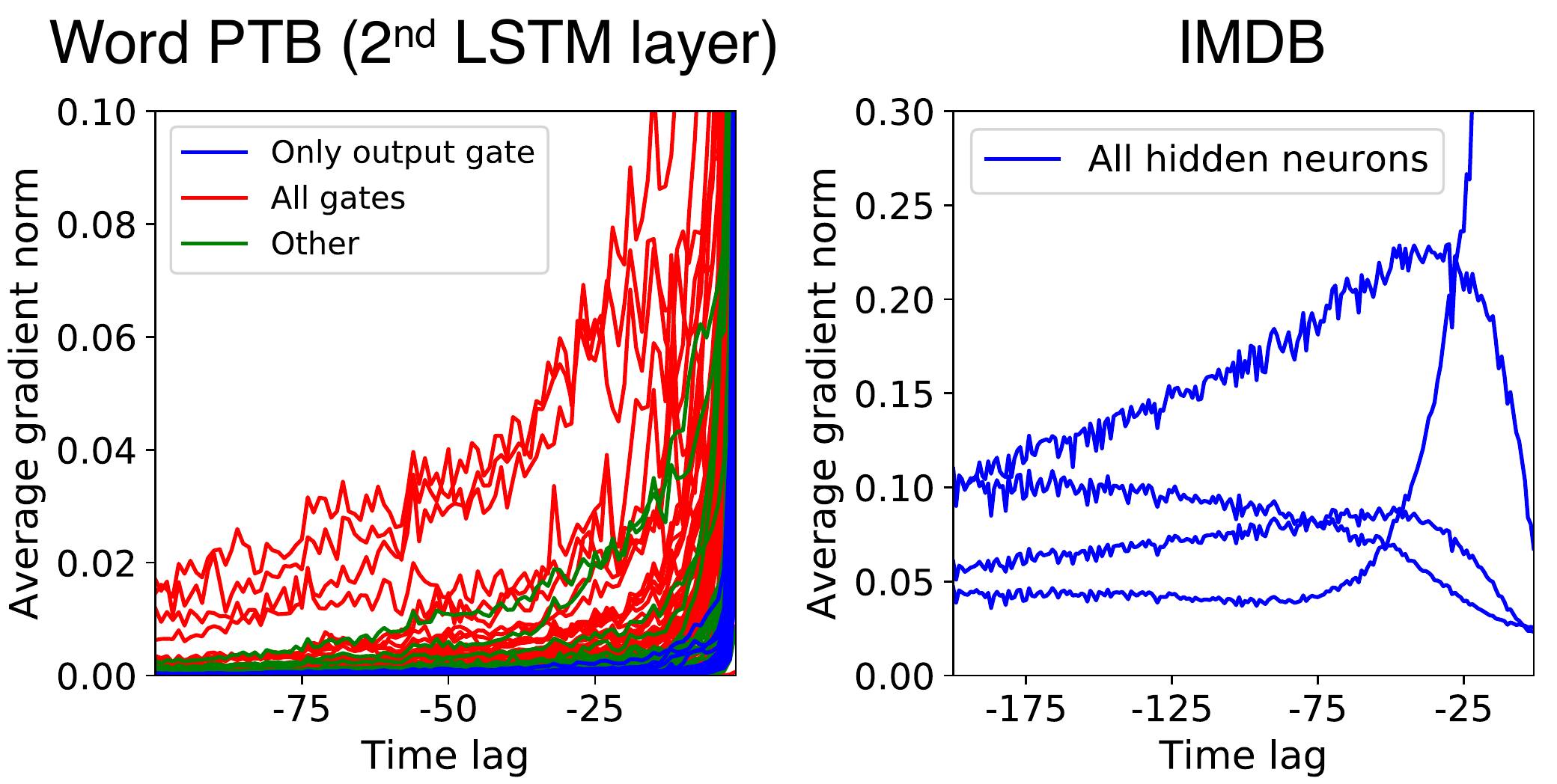}
\caption{Averaged over validation sequences, norm of the gradients of hidden neurons w.\,r.\,t.\ LSTM input for different time-lag for Bayes\,W+G+N.}
\label{fig:grads}
\end{wrapfigure}
\noindent character-level LM task is more challenging than the word level one: the model uses the whole gate mechanism to solve it. We think this is the main reason why gate sparsification does not help here.

As can be seen in fig.~\ref{fig:idea}, right, in the second LSTM layer of the small word-level language model, a lot of neurons have only one non-constant gate --- output gate. We investigate the described effect and find that the neurons with only non-constant output gate learn short-term dependencies while neurons with all non-constant gates usually learn long-term dependencies. 
To show that, we compute the gradients of each hidden neuron of the second LSTM layer w.\,r.\,t.\ the input of this layer at different lag $t$ and average the norm of this gradient over the validation set (see fig.~\ref{fig:grads}). 
The neurons with only non-constant output gate are ``short'': the gradient is large only for the latest timesteps and small for old timesteps. On the contrary, neurons with all non-constant gates
are mostly ``long'': the gradient is non-zero even for old timesteps. In other words, changing input 20--100 steps ago does not affect ``short''  neurons too much, which is not true for the ``long' neurons.
The presence of such  ``short'' neurons is expectable for the language model: neurons without memory quickly adapt to the latest changes in the input sequence and produce relevant output. 

In fact, for the neurons with only non-constant output gate, the memory cell $c_t$ is either monotonically increasing or monotonically decreasing depending on the sign of constant information flow $g$ so $tanh(c_t)$ always equals either to $-1$ or +1\footnote{Except for the first few epochs because $c_t$ is initialized with 0 value.} and $h_t = o_t$ or $-o_t$. This means these neurons are simplified to vanilla recurrent units. 

For classification tasks, memorizing information about the whole input sequence until the last timestep is important, therefore information flow $g$ is non-constant and saves information from the input to the memory. In other words, long dependencies are highly important for the classification. Gradient plots (fig.~\ref{fig:grads}) confirm this claim: the values of the neurons are strongly influenced by both old and latest inputs. Gradients are bigger for the short lag only for one neuron because this neuron focuses not only on the previous hidden states but also on reading the current inputs.

{\bf Gate structure intrinsically exists in LSTM.} 
As discussed above, the most visible gate structures are obtained for IMDB classification (a lot of constant output gates and non-constant information flow) and for the second LSTM layer of the small word-level LM task (a lot of neurons with only non-constant output gates). 
In our experiments, for these tasks, the same gate structures are detected even with unstructured sparsification, but with lower overall compression and less number of constant gates, see Appendix~\ref{unstructured}. This shows that the gate structure intrinsically exists in LSTM and depends on the task. The proposed method utilizes this structure to achieve better compression.

\begin{figure*}[hb]
    \centering
        \begin{tabular}{ll}
           \includegraphics[height=1.6cm]{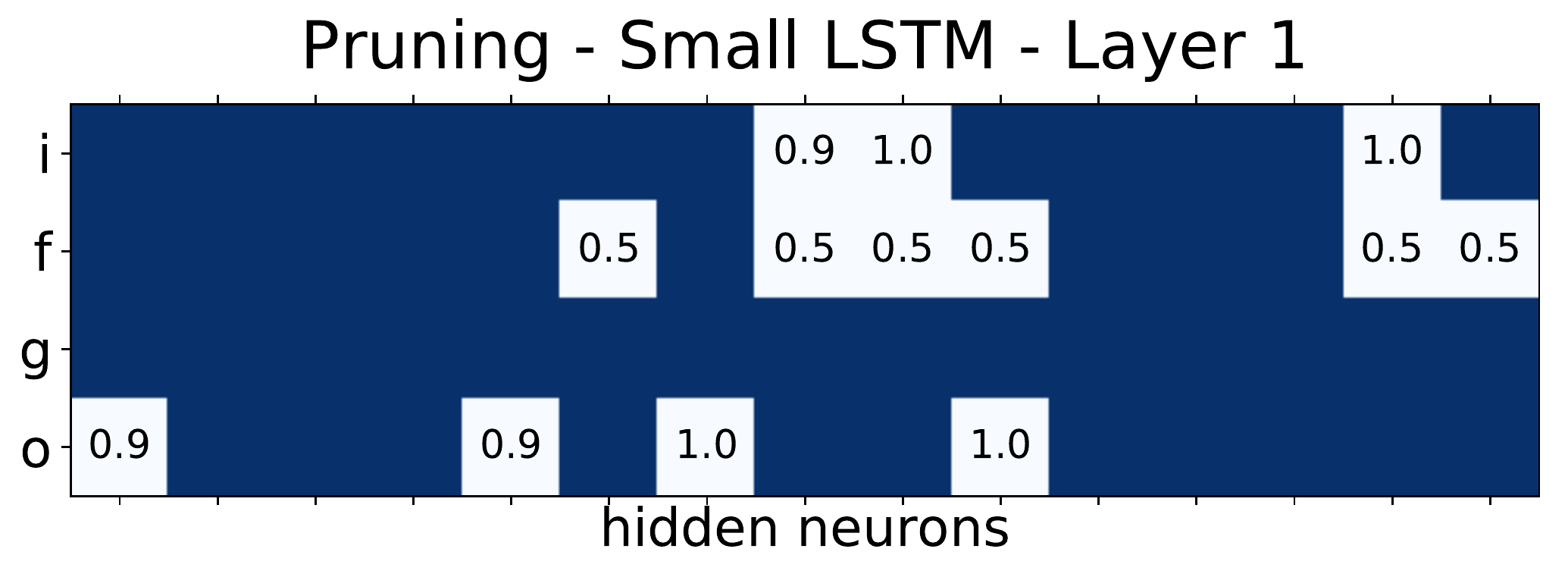} 
           & \includegraphics[height=1.6cm]{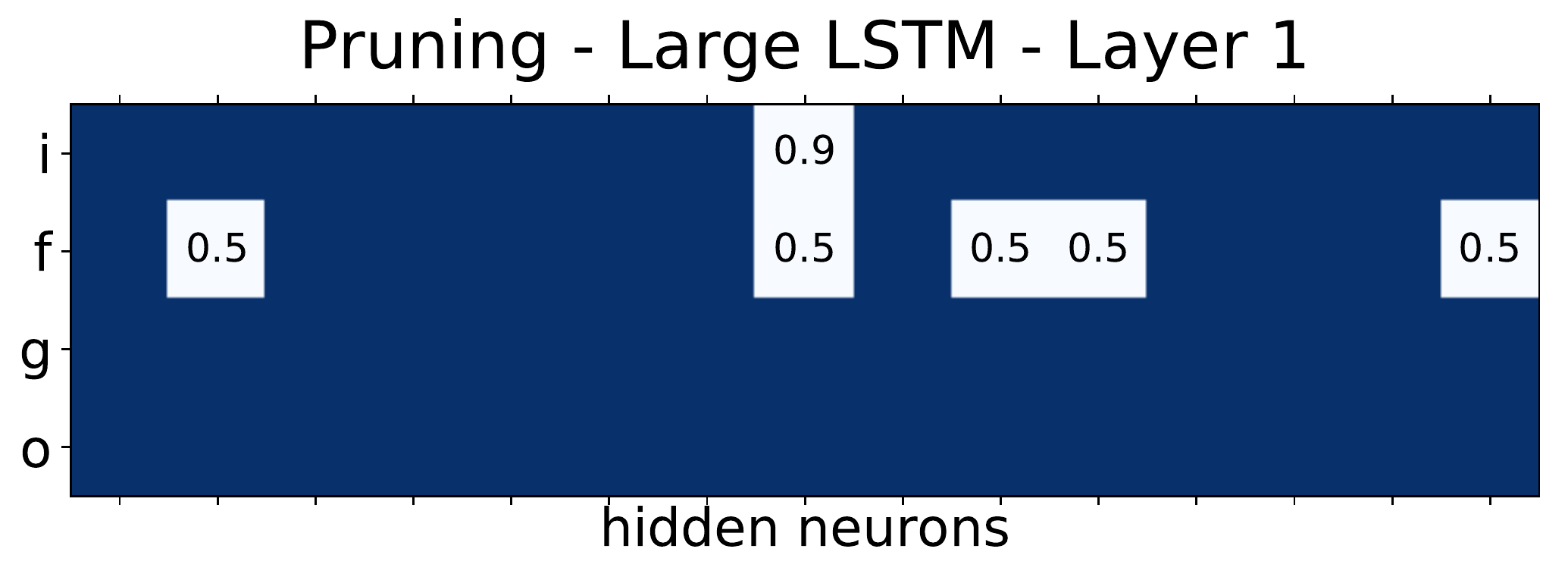} \\
           \includegraphics[height=1.6cm]{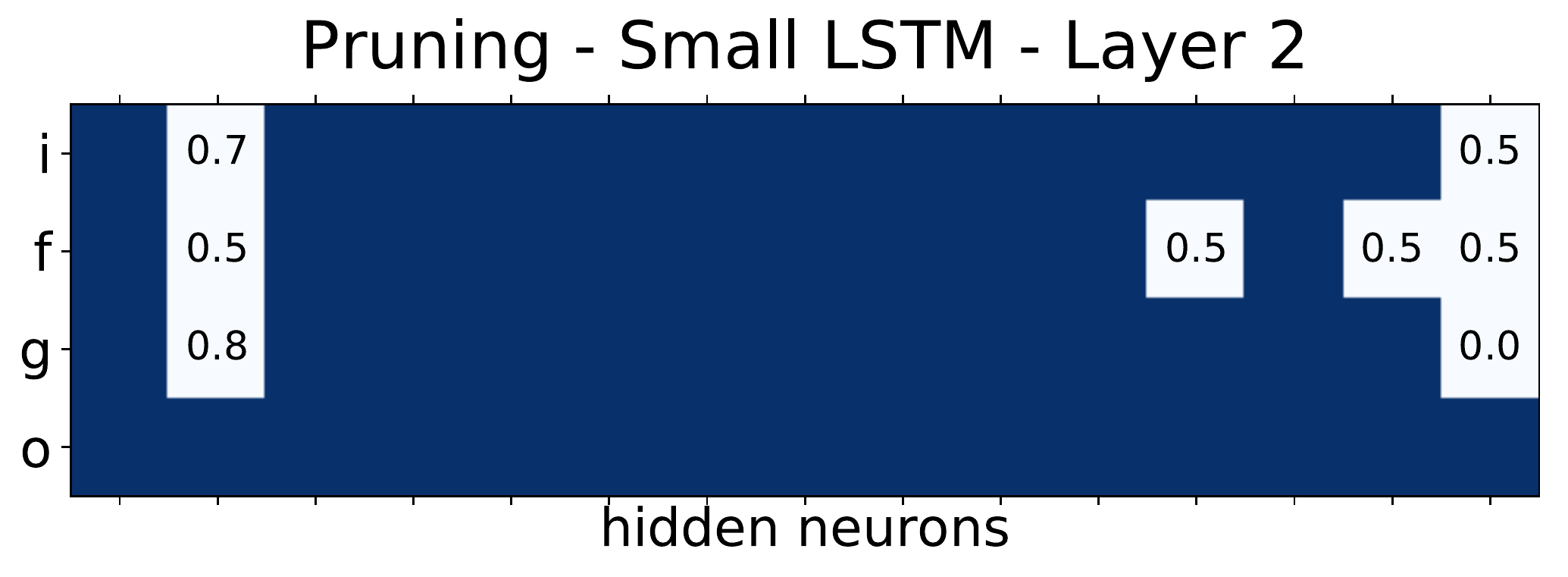} 
           & \includegraphics[height=1.6cm]{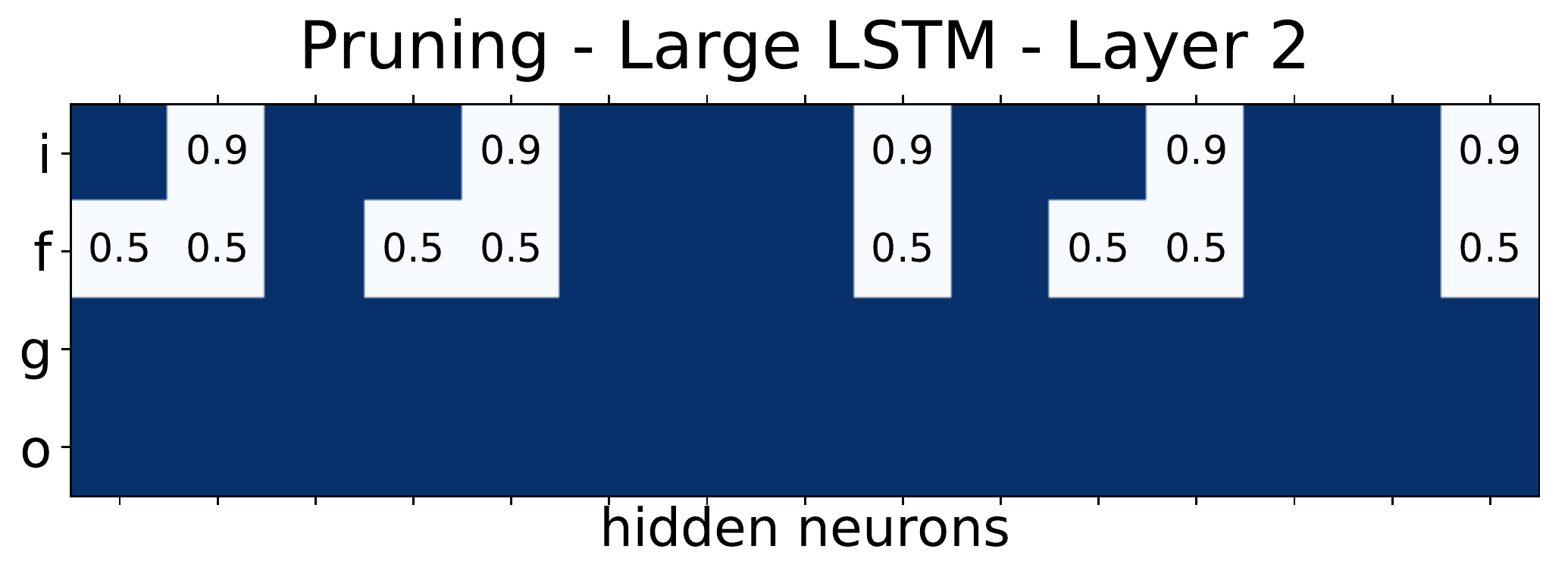} \\
        \end{tabular}
        \caption{Gate structure for word-level LM for Pruning W+G+N for different model sizes. 
        }
        \label{fig:gatesss}
\end{figure*}

We obtain a similar effect when we compare gate structures for the small word-level LM obtained using two different sparsification techniques: Bayes W+G+N (fig. \ref{fig:idea}, right) and Pruning W+G+N (fig. \ref{fig:gatesss}, left). The same gates become constant in these models. For the large language model (fig.~\ref{fig:gatesss}, right), the structure is slightly different than for the small model. It is expected because there is a significant  quality gap between these two models, so their intrinsic structure may be different.

\section*{Acknowledgments}
This research is in part based on the work supported by Samsung Research, Samsung Electronics.

\bibliographystyle{apalike}
\bibliography{bibliography.bib}

\begin{thebibliography}{}

\bibitem[Chirkova et~al., 2018]{emnlp}
Chirkova, N., Lobacheva, E., and Vetrov, D. (2018).
\newblock Bayesian compression for natural language processing.
\newblock In {\em Proceedings of the Conference on Empirical Methods in Natural
  Language Processing (EMNLP)}.

\bibitem[Cho et~al., 2014]{gru}
Cho, K., van Merri{\"{e}}nboer, B., G{\"{u}}l{\c c}ehre, {\c C}., Bahdanau, D.,
  Bougares, F., Schwenk, H., and Bengio, Y. (2014).
\newblock Learning phrase representations using rnn encoder--decoder for
  statistical machine translation.
\newblock In {\em Proceedings of the 2014 Conference on Empirical Methods in
  Natural Language Processing (EMNLP)}, pages 1724--1734, Doha, Qatar.
  Association for Computational Linguistics.

\bibitem[Hochreiter and Schmidhuber, 1997]{lstm}
Hochreiter, S. and Schmidhuber, J. (1997).
\newblock Long short-term memory.
\newblock {\em Neural Comput.}, 9(8):1735--1780.

\bibitem[Kingma and Ba, 2015]{adam}
Kingma, D.~P. and Ba, J. (2015).
\newblock Adam: {A} method for stochastic optimization.
\newblock In {\em Proceedings of the 3rd International Conference for Learning
  Representations (ICLR)}.

\bibitem[Louizos et~al., 2017]{chris}
Louizos, C., Ullrich, K., and Welling, M. (2017).
\newblock Bayesian compression for deep learning.
\newblock In {\em Advances in Neural Information Processing Systems 30}, pages
  3288--3298.

\bibitem[Maas et~al., 2011]{IMDB}
Maas, A.~L., Daly, R.~E., Pham, P.~T., Huang, D., Ng, A.~Y., and Potts, C.
  (2011).
\newblock Learning word vectors for sentiment analysis.
\newblock In {\em Proceedings of the 49th Annual Meeting of the Association for
  Computational Linguistics: Human Language Technologies - Volume 1}, pages
  142--150.

\bibitem[Marcus et~al., 1993]{ptb}
Marcus, M.~P., Marcinkiewicz, M.~A., and Santorini, B. (1993).
\newblock Building a large annotated corpus of english: The penn treebank.
\newblock {\em Comput. Linguist.}, 19(2):313--330.

\bibitem[Mikolov et~al., 2011]{mikolov11}
Mikolov, T., Kombrink, S., Burget, L., Cernocky, J., and Khudanpur, S. (2011).
\newblock Extensions of recurrent neural network language model.
\newblock In {\em 2011 IEEE International Conference on Acoustics, Speech and
  Signal Processing (ICASSP)}, pages 5528--5531.

\bibitem[Mikolov et~al., 2013]{NIPS2013_5021}
Mikolov, T., Sutskever, I., Chen, K., Corrado, G.~S., and Dean, J. (2013).
\newblock Distributed representations of words and phrases and their
  compositionality.
\newblock In {\em Advances in Neural Information Processing Systems 26}, pages
  3111--3119.

\bibitem[Molchanov et~al., 2017]{dmolch}
Molchanov, D., Ashukha, A., and Vetrov, D. (2017).
\newblock Variational dropout sparsifies deep neural networks.
\newblock In {\em Proceedings of the 34th International Conference on Machine
  Learning, ICML 2017}.

\bibitem[Narang et~al., 2017]{pruning}
Narang, S., Diamos, G.~F., Sengupta, S., and Elsen, E. (2017).
\newblock Exploring sparsity in recurrent neural networks.
\newblock In {\em Proceedings of the International Conference for Learning
  Representations (ICLR)}.

\bibitem[Pennington et~al., 2014]{pennington2014glove}
Pennington, J., Socher, R., and Manning, C.~D. (2014).
\newblock Glove: Global vectors for word representation.
\newblock In {\em Proceedings of the Conference on Empirical Methods in Natural
  Language Processing}, volume~14, pages 1532--1543.

\bibitem[See et~al., 2016]{intel}
See, A., Luong, M.-T., and Manning, C.~D. (2016).
\newblock Compression of neural machine translation models via pruning.
\newblock In {\em Proceedings of The 20th SIGNLL Conference on Computational
  Natural Language Learning}, pages 291--301. Association for Computational
  Linguistics.

\bibitem[Wen et~al., 2018]{groupsparseLSTM}
Wen, W., He, Y., Rajbhandari, S., Zhang, M., Wang, W., Liu, F., Hu, B., Chen,
  Y., and Li, H. (2018).
\newblock Learning intrinsic sparse structures within long short-term memory.
\newblock In {\em International Conference on Learning Representations}.

\bibitem[Yuan and Lin, 2006]{group_lasso}
Yuan, M. and Lin, Y. (2006).
\newblock Model selection and estimation in regression with grouped variables.
\newblock {\em JOURNAL OF THE ROYAL STATISTICAL SOCIETY, SERIES B}, 68:49--67.

\bibitem[Zaremba et~al., 2014]{zaremba14}
Zaremba, W., Sutskever, I., and Vinyals, O. (2014).
\newblock Recurrent neural network regularization.
\newblock In {\em arXiv preprint arXiv:1409.2329}.

\bibitem[Zhang et~al., 2015]{agnews}
Zhang, X., Zhao, J., and LeCun, Y. (2015).
\newblock Character-level convolutional networks for text classification.
\newblock In {\em Advances in Neural Information Processing Systems 28: Annual
  Conference on Neural Information Processing Systems (NIPS)}.

\end{thebibliography}

\appendix
\section{Technical details on the implementation of the idea in pruning\label{A}}
Consider a dataset of $N$ sequences $(x_i, y_i)$ and 
a model $p(y|x, W, b)$ defined by a recurrent neural network with weights $W$ and biases $b$.

To implement our idea about three levels of sparsification, for each neuron $\eta$, we define five (intersecting) sets of weights $ w_{\eta, i},\, w_{\eta, f},\,  w_{\eta, g},\,  w_{\eta, o},\,  w_{\eta, h}$.
The first four sets of weights correspond to four gates (dotted horizontal lines in~fig. \ref{fig:words}), and the last set corresponds to the neuron (solid vertical lines in fig.~\ref{fig:words}).
We apply group Lasso regularization~\cite{group_lasso} to these groups. We also apply Lasso regularization to the individual weights.

Following~\cite{groupsparseLSTM}, we set to zero all the individual weights with absolute value less than the threshold. If for some $\eta$ all the weights in $ w_{\eta, h}$ are set to zero, we remove the corresponding neuron as it does not affect the network's output.
If for some gate (for example, $f$) all the weights in $w_{\eta, f}$ are set to zero, we mark this gate as constant.

In contrast to our approach, in~\cite{groupsparseLSTM}, group Lasso is applied to larger groups $w_\eta$:
\[
w_\eta = w_{\eta, i} \cup w_{\eta, f} \cup w_{\eta, g} \cup w_{\eta, o} \cup w_{\eta, h}.
\]
They eliminate a neuron $\eta$ if all the weights in $w_\eta$ are zero. This approach does not lead to the sparse gate structure.

\section{Technical details on the implementation of the idea in Bayesian framework\label{B}}
{\bf Sparse variational dropout.}
Our approach relies on Sparse variational dropout~\cite{dmolch} (SparseVD). This model treats the weights of the neural network as random variables and comprises a log-uniform prior over the weights: $p(|w_{ij}|) \propto \frac{1}{|w_{ij}|}$ and a fully factorized normal approximate posterior over the weights: $q(w_{ij}) = \mathcal{N}(w_{ij}| m_{ij}, \sigma^2_{ij})$. Biases are treated as deterministic parameters. 
To find the parameters of the approximate posterior distribution and biases, the evidence lower bound (ELBO) is optimized:
\begin{equation}
\label{elbo}
\sum_{i=1}^N \mathbb{E}_{q(W|m, \sigma)} \log p(y^i|x^i, W, b)
- KL(q(W|m, \sigma)||p(W)) \rightarrow \max_{m, \sigma, b}
\end{equation}
Because of the log-uniform prior, for the majority of weights, the signal-to-noise ratio 
$m^2_{ij}/\sigma^2_{ij} \rightarrow 0$ and these weights do not affect the network's output. 
In~\cite{emnlp}, SparseVD is adapted to the RNNs.

{\bf Our model.}
To sparsify the individual weights, we apply SparseVD~\cite{dmolch} to all the weights $W$ of the LSTM, taking into account the recurrent specifics underlined in~\cite{emnlp}. To compress the layer and remove the hidden neurons, we follow~\cite{chris} and introduce group weights $z^h$ for the hidden neurons of the LSTM. 

The key component of our model is introducing groups weights $z^i,\, z^f, \, z^g, \, z^o$ on the preactivations of the gates and information flow. The resulting LSTM layer looks as follows:
\begin{align}
 f = \sigma \biggl( \bigl(W^x_f x_t + W^h_f h_{t-1} \bigr)  \odot z^f  +b_f \biggr)
 \quad \text{\{same for $i$, $o$ and $g$\}} \nonumber \\
 c_t = f \odot c_{t-1} +  i \odot  g \quad  
~~~h_t =  o \odot tanh(c_t) \odot z^h  \nonumber
\end{align}
The model is equivalent to multiplying the rows and columns of the weight matrices by the group weights:
\[
\hat w^h_{f, ij} = w^h_{f, ij} ~\cdot z^h_{i} ~\cdot z^f_{j}
\quad \{\text{same for } i, o \text{ and } g \}
\]

If some component of $z^i$, $z^f$, $z^o$, or $z^g$ is set to zero, we mark the corresponding gate as constant.
If some component of $z^h$ is set to zero, we remove the corresponding neuron from the model.

{\bf Training our model.}
We work with the group weights $z$ in the same way as with the weights $W$: we approximate the posterior with the fully factorized normal distribution given the fully factorized log-uniform prior distribution.
To estimate the expectation in~\eqref{elbo}, we sample weights from the approximate posterior distribution in the same way as in~\cite{emnlp}.

With the integral estimated with one Monte-Carlo sample, the first term in~\eqref{elbo} becomes the usual loss function (for example, cross-entropy in language modeling). The second term is a regularizer depending on the parameters $\mu$ and $\sigma$ (for the exact formula, see~\cite{dmolch}).

After learning, we zero out all the weights and the group weights with the signal-to-noise ratio less than 0.05. At the testing stage, we use the mean values of all the weights and the group weights.

\section{Experimental setup} \label{setup}

{\bf Datasets.} To evaluate our approach on the text classification task, we use two standard datasets: IMDb dataset~\cite{IMDB} for binary classification and AGNews dataset~\cite{agnews} for four-class classification. We set aside 15\% and 5\% of the training data for validation purposes respectively. For both datasets, we use a vocabulary of 20,000 most frequent words. To evaluate our approach on the language modeling task, we use the Penn Treebank corpus~\cite{ptb} with the train/valid/test partition from~\cite{mikolov11}. The dataset has a vocabulary of 50 characters or 10,000 words.

\paragraph{Pruning.}
All the small models including baseline are trained without dropout as in standard TensorFlow implementation. We train them from scratch for 20 epochs with SGD with a decaying learning rate schedule: an initial learning rate is equal to $1$, the learning rate starts to decay after the $4$-th epoch, the learning rate decay is equal to $0.6$. For the two-level sparsification (W+N), we use Lasso regularization with $\lambda=1e-5$ and group Lasso regularization with $\lambda=0.002$. For the three-level sparsification (W+G+N), we use Lasso regularization with $\lambda=1e-5$ and group Lasso regularization with $\lambda=0.0017$. We use the threshold $1e-4$ to prune the weights in both models during training.

All the large models including baseline are trained in the same setting as in~\cite{groupsparseLSTM} except for the group Lasso regularization because we change the weight groups. We use the code provided by the authors. Particularly, we use binary dropout~\cite{zaremba14} with the same dropout rates. We train the models from scratch for 55 epochs with SGD with a decaying learning rate schedule: an initial learning rate is equal to $1$, the learning rate decreases two times during training (after epochs 18 and 36), the learning rate decay is equal to $0.2$ and $0.1$ for two- and three-level sparsification correspondingly. For the two-level sparsification (W+N), we use Lasso regularization with $\lambda=1e-5$ and group Lasso regularization with $\lambda=0.0015$. For the three-level sparsification (W+G+N), we use Lasso regularization with $\lambda=1.5e-05$ and group Lasso regularization with $\lambda=0.00125$. We use the same threshold $1e-4$ as in the small models.

\paragraph{Bayesian sparsification.}
In all the Bayesian models, we sparsify the weight matrices of all layers. Since in text classification tasks, usually only a small number of input words are important, we use additional multiplicative weights to sparsify the input vocabulary following~\citet{emnlp}. For the networks with the embedding layer, in configurations W+N and W+G+N, we also sparsify the embedding components (by introducing group weights $z^x$ multiplied by $x_{t}$.) 

We train our networks using Adam~\cite{adam}. Baseline networks overfit for all our tasks, therefore, we  present results for them with early stopping. Models for the text classification and the character-level LM are trained in the same setting as in~\cite{emnlp} (we used the code provided by the authors). For the text classification tasks, we use a learning rate equal to $0.0005$ and train Bayesian models for 800 / 150 epochs on IMDb / AGNews. The embedding layer for IMDb / AGNews is initialized with word2vec~\cite{NIPS2013_5021} / GloVe~\cite{pennington2014glove}. For the language modeling tasks, we train Bayesian models for 250 / 50 epochs on character-level / word-level tasks using a learning rate of $0.002$. 

For all the weights that we sparsify, we initialize $\log\sigma$ with -3. We eliminate weights with the signal-to-noise  ratio less than $\tau=0.05$. To compute the number of the remaining neurons or non-constant gates, we use the corresponding rows/columns of $W$ and the corresponding weights $z$ if applicable. 

\section{Experiments with unstructured Bayesian sparsification} \label{unstructured}
\begin{figure}[H]
    \centering
         \includegraphics[height=1.6cm]{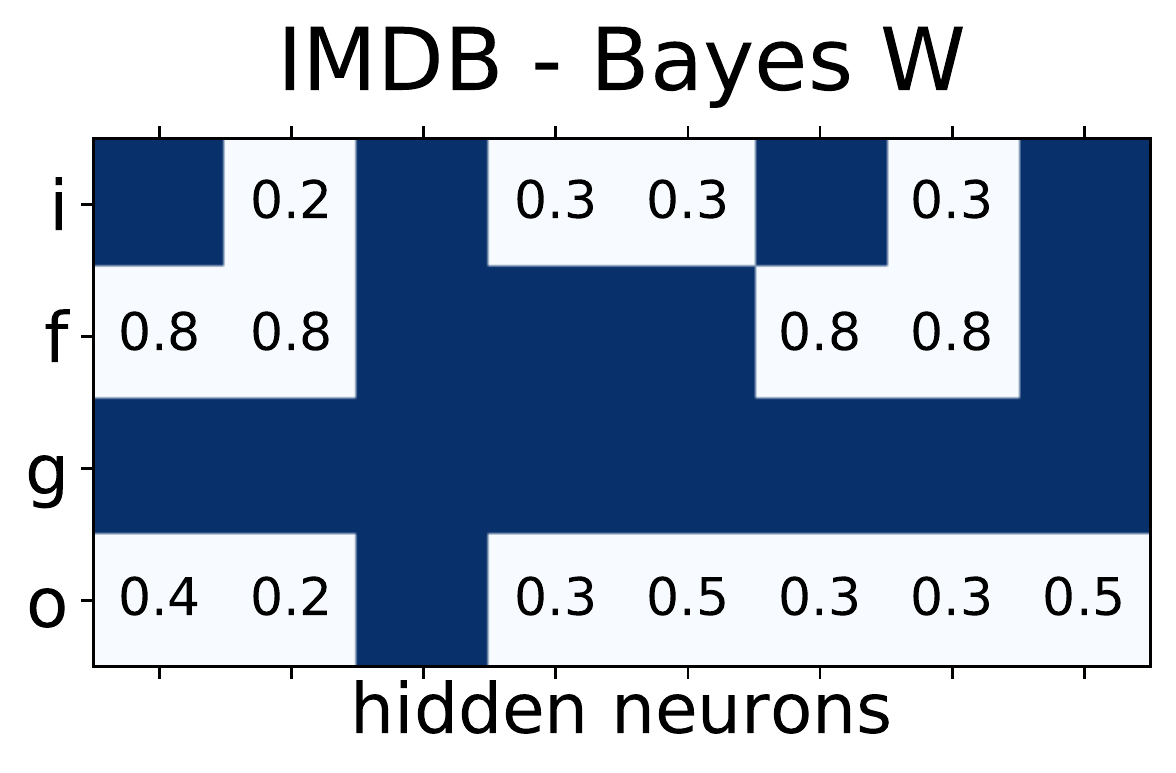} \hspace{0.5cm}
           \includegraphics[height=1.6cm]{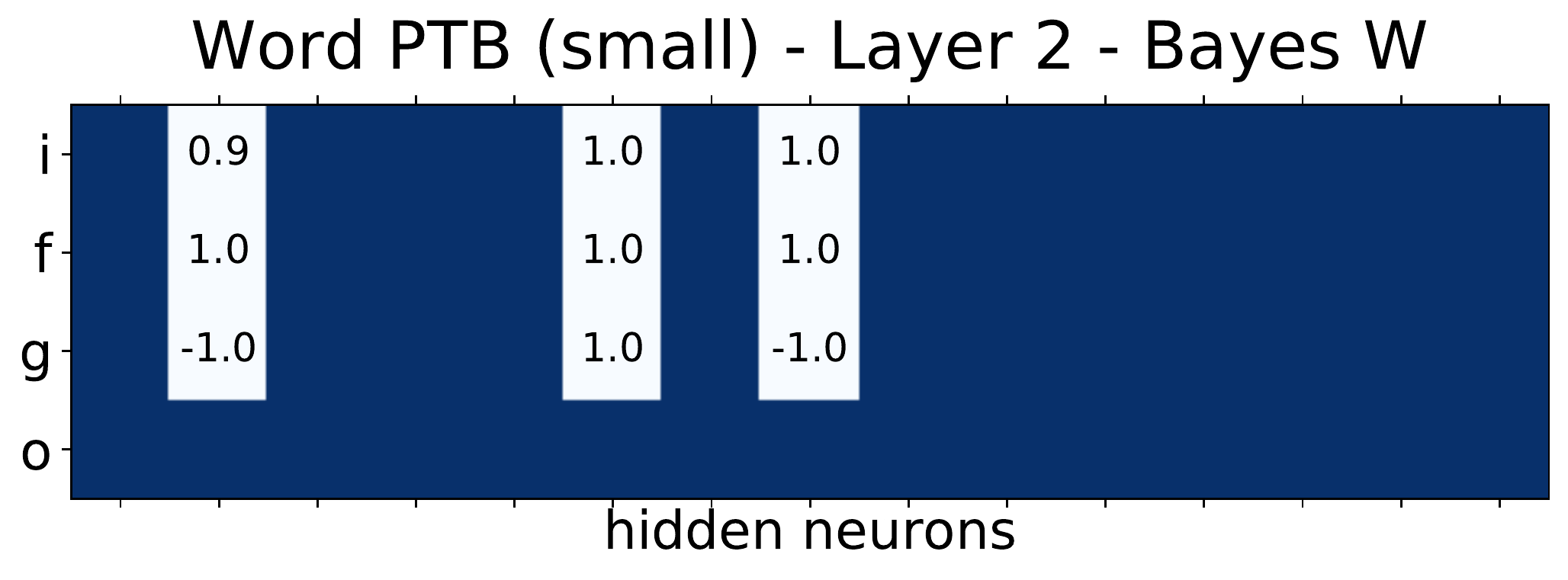} 
        \caption{Structure of the gate sparsity for the text classification and word-level language modeling obtained with Bayes W. Constant gates are shown in white with the corresponding activation values. For language modeling, only 15 randomly chosen active neurons are presented.}
        \label{fig:unstructured}
\end{figure}

In this section, we present experimental results for the unstructured Bayesian sparsification (configuration Bayes W). This configuration corresponds to a model of~\citet{emnlp}. Table~\ref{tab:unstructured} shows quantitative results, and figure~\ref{fig:unstructured} shows the resulting gate structures for the IMDB classification task and the second LSTM layer of the word-level language modeling task. Since Bayes W model does not comprise any group weights, the overall compression of the RNN is lower than for Bayes W+G+N (tab.~\ref{tab:pruning}), so there are more non-constant gates. However, the patterns in gate structures are the same as in Bayes W+G+N gate structures (fig.~\ref{fig:idea}): for the IMDB classification, the model has a lot of constant output gates and non-constant information flow, for language modeling, the model has neurons with only non-constant output gates. 

\begin{table*}
	\centering
	\begin{tabular}{clcccc}
		Task & Method & Quality & Compression  & Neurons & Gates\\ 
		\hline
		IMDb&Bayes W~\cite{emnlp}  & 83.62 & 18567 & $8$ & 17\\  
		\hline
		AGNews &Bayes W~\cite{emnlp}  & 89.14 & 561x &$ 34$& $ 76$ \\
		\hline
		Char PTB &Bayes W~\cite{emnlp} & $1.472 - 1.429$ &  7.9x  & $431$& 1718\\
		\hline
		Word PTB&Bayes W~\cite{emnlp} & 114.80 -- 109.85  & 10.52x & 55 -- 124 & 218 -- 415 \\
	\end{tabular}
	\caption{Quantitative results for the unstructured Bayesian sparsification.  Bayes W corresponds to the SparseVD method of~\citet{emnlp}. For language modeling, we evaluate quality on validation and test sets. Compression is equal to $|W|/|W\neq0|$.
		In  the last  columns, the numbers of the remaining hidden neurons and non-constant gates in the LSTM layers are reported.}\label{tab:unstructured}
\end{table*}

\end{document}